\documentclass[10pt,twocolumn,letterpaper]{article}

\usepackage{cvpr}              

\usepackage{graphicx}
\usepackage{amsmath}
\usepackage{amssymb}
\usepackage{booktabs}
\usepackage{makecell}
\usepackage{multirow}
\usepackage[pagebackref,breaklinks,colorlinks]{hyperref}

\usepackage[capitalize]{cleveref}
\crefname{section}{Sec.}{Secs.}
\Crefname{section}{Section}{Sections}
\Crefname{table}{Table}{Tables}
\crefname{table}{Tab.}{Tabs.}

\def\confName{CVPR}
\def\confYear{2023}

\begin{document}

\title{Semantic-Conditional Diffusion Networks for Image Captioning\thanks{{\small This work was performed at JD Explore Academy. The first two authors contributed equally to this work.}}}

\author{Jianjie Luo$^{\dag}$, Yehao Li$^{\ddag}$, Yingwei Pan$^{\ddag}$, Ting Yao$^{\ddag}$, Jianlin Feng$^{\dag}$, Hongyang Chao$^{\dag}$, and Tao Mei$^{\ddag}$\\
{\normalsize\centering$^{\dag}$ Sun Yat-sen University, Guangzhou, China}\\
{\normalsize\centering$^{\ddag}$ JD Explore Academy, Beijing, China}\\
{\tt\small \{jianjieluo.sysu,yehaoli.sysu,panyw.ustc,tingyao.ustc\}@gmail.com;}\\
{\tt\small\{fengjlin,isschhy\}@mail.sysu.edu.cn;tmei@jd.com}
}

\maketitle

\begin{abstract}
Recent advances on text-to-image generation have witnessed the rise of diffusion models which act as powerful generative models. Nevertheless, it is not trivial to exploit such latent variable models to capture the dependency among discrete words and meanwhile pursue complex visual-language alignment in image captioning. In this paper, we break the deeply rooted conventions in learning Transformer-based encoder-decoder, and propose a new diffusion model based paradigm tailored for image captioning, namely Semantic-Conditional Diffusion Networks (SCD-Net). Technically, for each input image, we first search the semantically relevant sentences via cross-modal retrieval model to convey the comprehensive semantic information. The rich semantics are further regarded as semantic prior to trigger the learning of Diffusion Transformer, which produces the output sentence in a diffusion process. In SCD-Net, multiple Diffusion Transformer structures are stacked to progressively strengthen the output sentence with better visional-language alignment and linguistical coherence in a cascaded manner. Furthermore, to stabilize the diffusion process, a new self-critical sequence training strategy is designed to guide the learning of SCD-Net with the knowledge of a standard autoregressive Transformer model. Extensive experiments on COCO dataset demonstrate the promising potential of using diffusion models in the challenging image captioning task. Source code is available at \url{https://github.com/YehLi/xmodaler/tree/master/configs/image_caption/scdnet}.
\end{abstract}

\section{Introduction}
\label{sec:intro}

\begin{figure}[!tb]
 \vspace{-0.3in}
    \centering {\includegraphics[width=0.439\textwidth]{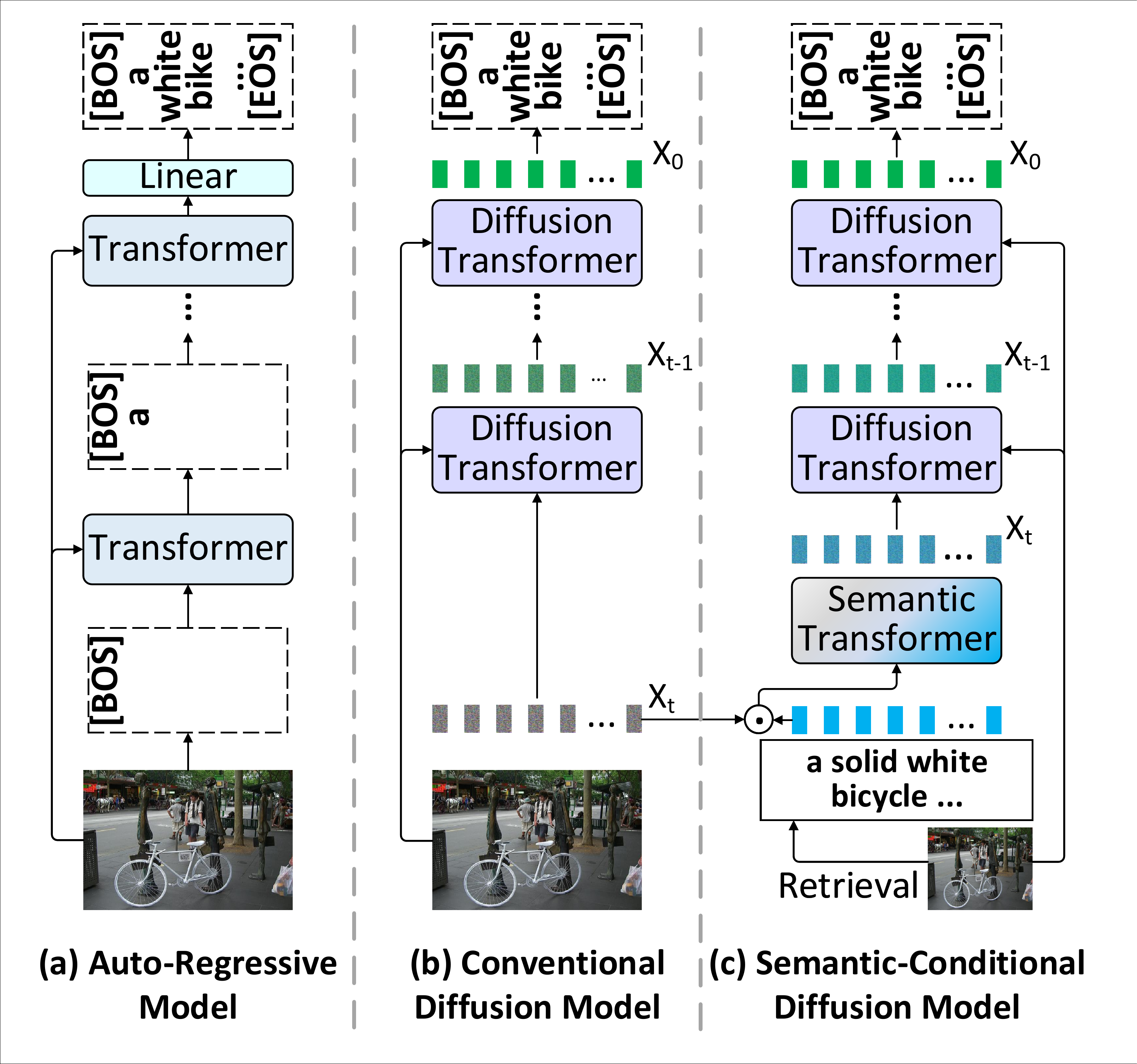}}
    \vspace{-0.15in}
    \caption{An illustration of (a) typical autoregressive model, (b) conventional diffusion model in \cite{analogbits}, and (c) our semantic-conditional diffusion model in SCD-Net.}
    \label{fig:intro}
    \vspace{-0.28in}
\end{figure}

As one of the fundamental tasks in vision and language field, image captioning aims to describe the interested semantics in a natural sentence.
This task naturally connects computer vision and natural language processing by perceiving visual content in a scene and interpreting them in human language, simulating a basic capability of human intelligence.
The dominate force in current state-of-the-art techniques \cite{sharma2018conceptual,herdade2019image,cornia2020meshed,li2022comprehending,yao2019hierarchy} is to capitalize on encode-decoder structure and frame the learning process in an autoregressive manner. In particular, an image encoder is responsible for encoding visual content as high-level semantics, and a sentence decoder learns to decode the sequential sentence word-by-word (Fig.\ref{fig:intro} (a)).
Nevertheless, such autoregressive progress only allows for unidirectional textual message passing, and typically relies on considerable computational resources that scale quadratically w.r.t. the sentence length.

To alleviate this limitation, recent advances \cite{nat2018,gao2019masked,liu2021o2na} start to emerge as a non-autoregressive solution that enables bidirectional textual message passing and emits all words in parallel, leading to a light-weight and scalable paradigm.
However, these non-autoregressive approaches are generally inferior to the autoregressive methods. The performance degradation can be mostly attributed to the word repetition or omissions problems if leaving the sequential dependency under-exploited. In addition, the indisposed sentence quality also makes it difficult to upgrade non-autoregressive solution with powerful self-critical sequence learning \cite{rennie2017self} widely adopted in autoregressive methods.

More recently, a superior generative module named diffusion model \cite{ddpm} has brought forward milestone improvements for visual content generation. This motivates a recent pioneering practice \cite{analogbits} to explore diffusion model for image captioning, pursuing non-autoregressive sentence generation in a high degree of parallelism. It manages to enable continuous diffusion process to produce the discrete word sequence by representing each word as binary bits. Different from the typical discrete sentence generation in a single shot, such continuous diffusion process (Fig.\ref{fig:intro} (b)) can be represented as a parameterized Markov chain that gradually adds Gaussian noise to the sentence. Each reverse state transition is thus learnt to recover the original sentence data from noise-augmented data via denoising.
Despite showing comparable performances against basic autoregressive models, the severe problem of word repetition or omissions is still overlooked.

In an effort to mitigate this problem, we devise a new diffusion model based non-autoregressive paradigm, called Semantic-Conditional Diffusion Networks (SCD-Net). Our launching point is to introduce the comprehensive semantic information of the input images into the continuous diffusion process, which act as semantic prior to guide the learning of each reverse state transition (Fig.\ref{fig:intro} (c)). SCD-Net is henceforth able to encourage a better semantic alignment between visual content and the output sentence, i.e., alleviating the omission of semantic words. As a by-product, our SCD-Net enables the powerful self-critical sequence learning during the continuous diffusion process. Such sentence-level optimization of diffusion process strengthens the linguistical coherence of the output sentence, and thus alleviates word repetition issue.

Technically, our SCD-Net is composed of cascaded Diffusion Transformer structures that progressively enhance the output sentences. Each Diffusion Transformer exploits the semantic-conditional diffusion process to learn Transformer-based encoder-decoder. In particular, for each input image, Diffusion Transformer first retrieves the semantically relevant words by using an off-the-shelf cross-modal
retrieval model. These semantic words are additionally incorporated into the continuous diffusion process, targeting for constraining reverse state transition with the semantic condition. More importantly, we upgrade the semantic-conditional diffusion process with a new guided self-critical sequence learning strategy. This strategy elegantly transfers the knowledge of a standard autoregressive Transformer model to our non-autoregressive Diffusion Transformer through sentence-level reward, leading to a stabilized and boosted diffusion process.

In sum, we have made the following contributions: (\textbf{I}) SCD-Net designs a novel semantic-conditional diffusion process for image captioning, pursuing a scalable non-autoregressive paradigm with better visual-language alignment.
(\textbf{II}) SCD-Net paves a new way to couple the continuous diffusion process with a new guided self-critical sequence learning.
(\textbf{III}) SCD-Net has been properly analyzed and verified through extensive experiments on COCO, demonstrating its encouraging potential.

\section{Related Work}
\subsection{Autoregressive Image Captioning}
Recent research on autoregressive image captioning has proceeded along two dimensions: RNN-based and Transformer-based approaches.

\textbf{RNN-based Approaches.} With the advent of deep learning \cite{imagenet,Bahdanau14,Sutskever:NIPS14}, modern image captioning techniques utilize CNN plus RNN to generate sentences, yielding flexible syntactical structure. For example, the earlier attempts \cite{karpathy2015deep,Vinyals14,donahue2015long} present an end-to-end neural network system that encodes the visual content with CNN and decodes the sentence with RNN. Attention mechanism \cite{Xu:ICML15} is further employed to dynamically focus on the relevant region when generating word at each decoding timestep. Meanwhile, semantic attributes \cite{You:CVPR16,yao2017boosting,yao2017incorporating,li2019pointing} are incorporated into RNN to complement visual representations for enhancing description generation. In contrast to traditional visual attention over equally-sized image regions, Up-Down \cite{anderson2017bottom} combines bottom-up and top-down attention mechanism to measure attention at object level. After that, GCN-LSTM \cite{yao2018exploring} models the relations between objects, which are injected into the top-down attention model to boost caption generation. SGAE \cite{Yang:CVPR19} utilizes scene graph structure to depict fine-grained semantics in the images, aiming to capture the language inductive bias.

\textbf{Transformer-based Approaches.} Recently, taking the inspiration from the success of Transformer \cite{vaswani2017attention} in NLP field, numerous Transformer-based image captioning models start to emerge. \cite{sharma2018conceptual} directly employs the primary Transformer structure in NLP on image captioning task. \cite{herdade2019image} incorporates the spatial relations among objects into Transformer. Meshed-Memory Transformer \cite{cornia2020meshed} improves both encoder and decoder in Transformer by learning prior knowledge with persistent memory and exploiting low- and high-level features with a mesh-like connections across multiple layers. X-Transformer \cite{pan2020x} further introduces higher order intra and inter-modal interactions to enhance the encoder and decoder of Transformer. Later on, APN \cite{yang2021auto} learns to parse trees unsupervisedly by imposing probabilistic graphical model on self-attention layers of Transformer for both captioning and visual question answering.

\subsection{Non-Autoregressive Image Captioning}
In contrast to autoregressive methods that generate sentences word-by-word, non-autoregressive methods emit all words in parallel, enabling bidirectional textual message passing. NAT \cite{nat2018} first proposes non-autoregressive decoding to improve the inference speed for neural machine translation, attracting increasing attention in captioning field. MNIC \cite{gao2019masked} presents masked non-autoregressive decoding to generate captions parallelly with enhanced semantics and diversity. O2NA \cite{liu2021o2na} first generates the object-oriented coarse-grained caption and then refines each object word to alleviate description ambiguity issue. SATIC \cite{zhou2021semi} proposes semi-autoregressive Transformer that predicts a group of words in parallel and generates the groups from left to right, pursuing a better trade-off between speed and quality. Most recently, sparked by the success of diffusion models \cite{ddpm} in image generation, Bit Diffusion \cite{analogbits} encodes the discrete text into binary bits, and utilizes diffusion model with self-conditioning for caption generation. Nevertheless, the performance of Bit Diffusion is still inferior to state-of-the-art autoregressive Transformer. Moreover, how to optimize the diffusion model with sentence-level reward (e.g., CIDEr \cite{vedantam2015cider}) has not yet been explored.

Our work also falls into the latter category of non-autoregressive image captioning with diffusion models. Our SCD-Net goes beyond Bit Diffusion by strengthening the visual-language semantic alignment with semantic-conditional diffusion process. A new guided self-critical sequence learning is designed to further stabilize and boost the diffusion process with sentence-level reward.

\section{Method}
\begin{figure*}[!tb]
\vspace{-0.0in}
    \centering {\includegraphics[width=1\textwidth]{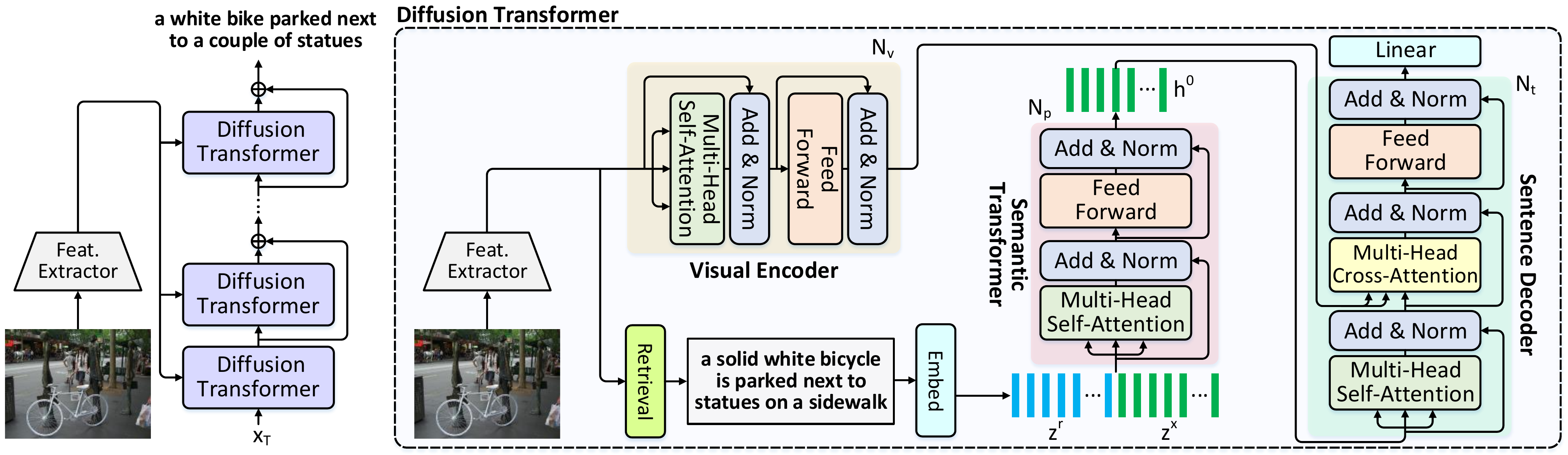}}
    \vspace{-0.2in}
    \caption{An overview of Semantic-Conditional Diffusion Networks (SCD-Net) consisting of multiple stacked Diffusion Transformers in a cascaded manner (left). Each Diffusion Transformer is composed of Visual Encoder, Semantic Transformer, and Sentence Decoder (right).}
    \label{fig:framework}
    \vspace{-0.2in}
\end{figure*}

In this section, we present our proposed Semantic-Conditional Diffusion Networks (SCD-Net), that facilitates diffusion process based image captioning with rich semantic prior. Figure \ref{fig:framework} depicts the cascaded framework of SCD-Net with multiple stacked Diffusion Transformers, and the detailed architecture of each Diffusion Transformer.

\subsection{Problem Formulation}

\textbf{Notation of Diffusion Model.}
Suppose we have an input image with $K$ interested objects, which will be described by a textual sentence $S$ in $N_s$ words. Let $\mathcal{V}=\{ {v_i}\} _{i = 1}^K$ denote the set of objects detected by Faster R-CNN \cite{anderson2017bottom,ren2015faster}, where $v_i \in \mathbb{R}^{D_v}$ represents the $D_v$-dimensional feature of each object. Here we basically formulate the procedure of caption generation with diffusion model \cite{ddpm,analogbits}. Considering that the words in the textual sentence are discrete data, we follow Bit Diffusion \cite{analogbits} and convert each word into $ n = \left\lceil {{{\log }_2}\mathcal{W}} \right\rceil $ binary bits (i.e., ${\{  0,1\} ^n}$), where $\mathcal{W}$ is the vocabulary size. In this way, the textual sentence $S$ is transformed into real numbers $x_0 \in \mathbb{R}^{n \times {N_s}}$ to trigger diffusion model. In particular, the diffusion model consists of two processes, i.e., the forward and reverse process.

\textbf{Forward Process.} The forward process is defined as a Markov chain that gradually adds Gaussian noise to the sentence data $x_0$. For any $t \in \left( {0,T} \right]$, the forward state transition \cite{vdm} from $x_0$ to $x_t$ is calculated as:
\begin{equation}
{x_t} = \sqrt {{\bf{sigmoid}}( - \gamma (t'))} {x_0} + \sqrt {{\bf{sigmoid}}(\gamma (t'))} \epsilon,
\end{equation}
where $t'=t/T$, $\epsilon \sim \mathcal{N}(0,\textit{I})$, $t \sim \mathcal{U}(0,T)$, $\mathcal{N}$ is normal distribution, $\mathcal{U}$ is uniform distribution, and $\gamma(t')$ is a monotonically increasing function. After that, a Diffusion Transformer $f(x_t, \gamma (t'), \mathcal{V})$ is trained to reconstruct $x_0$ conditioned on $\mathcal{V}$ through denoising in $\ell _2$ regression loss:
\begin{equation}
\mathcal{L}_{bit} = {\mathbb{E}_{t \sim \mathcal{U}(0,T),\epsilon  \sim \mathcal{N}(0,\textit{I})}}{\left\| {f({x_t},\gamma (t'),\mathcal{V}) - {x_0}} \right\|^2}.
\label{eq:diff}
\end{equation}

\textbf{Reverse Process.}
In an effort to generate sentences from the learnt Diffusion Transformer based on the given image, the reverse process samples a sequence of latent states $x_t$ from $t = T$ to $t = 0$. Specifically, given the number of steps $T$, we discretize time uniformly with width $1/T$ to obtain $s=t-1-\Delta$, $t'=t/T$, and $s'=s/T$. Next the reverse state transition $x_{t-1}$ is measured as:
\begin{equation}
\begin{aligned}
&{{\alpha _s} = \sqrt {{\bf{sigmoid}}( - \gamma (s'))} ,{\alpha _t} = \sqrt {{\bf{sigmoid}}( - \gamma (t'))} },\\
&{{\sigma _s} = \sqrt {{\bf{sigmoid}}(\gamma (s'))} },\\
&{c =  - {\bf{expm1}}(\gamma (s') - \gamma (t'))},\\
&{u(x{}_t;s',t') = {\alpha _s}({x_t}(1 - c)/{\alpha _t} + cf({x_t},\gamma (t'),\mathcal{V}))},\\
&{{\sigma ^2}(s',t') = \sigma _s^2c},\\
&{{x_{t - 1}} = u(x{}_t;s',t') + \sigma (s',t')\epsilon},
\end{aligned}
\end{equation}
where $\Delta$ denotes the time difference and ${\bf{expm1}}( \cdot )={\bf{exp}}( \cdot )-1$. After iteratively triggering Diffusion Transformer from $x_T$, we can obtain the estimation and a quantization operation is operated to transform it into bits $x_0$.

\subsection{Diffusion Transformer}
The basic Diffusion Transformer is constructed as a typical Transformer-based encoder-decoder structure, including a visual encoder and a sentence decoder. Formally, given the detected objects $\mathcal{V}$ in image, the visual encoder transforms them into visual tokens through self-attention. Then the visual tokens and the word tokens ${x_t} = \{ w_0^t,w_1^t,...,w_{N_s}^t \}$ at time step $t$ are fed into the sentence decoder for caption generation.

\textbf{Visual Encoder.} The visual encoder consists of $N_v$ stacked Transformer encoder blocks. Each block is composed of a multi-head self-attention layer plus a feed-forward layer. The $i$-th Transformer encoder block in the visual encoder operates as:
\begin{equation}
\begin{aligned}
&{{\mathcal{V}^{i + 1}} = {\bf{FFN}}({\bf{norm}}({\mathcal{V}^i} + {\bf{MultiHead}}({\mathcal{V}^i},{\mathcal{V}^i},{\mathcal{V}^i})))},\\
&{{\bf{FFN}}({\bf{Z}}) = {\bf{norm}}({\bf{Z}} + {\bf{FC}}(\delta ({\bf{FC}}({\bf{Z}}))))},\\
&{{\bf{MultiHead}}({\bf{Q}},{\bf{K}},{\bf{V}}) = {\bf{Concat}}({head}_{1},...,{head}_{H}){W^O}},\\
&{{head}_{i} = {\bf{Attention}}({\bf{Q}}W_i^Q,{\bf{K}}W_i^K,{\bf{V}}W_i^V)},\\
&{{\bf{Attention}}({\bf{Q}},{\bf{K}},{\bf{V}}) = {\bf{softmax}}(\frac{{{\bf{Q}}{{\bf{K}}^T}}}{{\sqrt d }}){\bf{V}}},
\end{aligned}
\end{equation}
where ${\bf{FFN}}$ is the feed-forward layer, ${\bf{MultiHead}}$ is the multi-head self-attention layer, $\bf{norm}$ is layer normalization, ${\bf{FC}}$ is the fully-connected layer, ${\bf{Concat}}( \cdot )$ is the concatenation operation, $\delta$ is the activation function, $W_i^Q$, $W_i^K$, $W_i^V$, $W^O$ are weight matrices of the $i$-th head, $H$ is the number of head, and $d$ is the dimension of each head. Note that the input of the first Transformer encoder block is the primary set of detected objects $\mathcal{V}^0=\mathcal{V}$. Accordingly, after stacking $N_v$ blocks, we achieve the contextually enhanced visual tokens ${\mathcal{\tilde V}}={\mathcal{V}^{N_v}}$.

\textbf{Sentence Decoder.} The sentence decoder contains $N_t$ stacked Transformer decoder blocks. Each block consists of one multi-head self-attention layer, one multi-head cross-attention layer, and one feed-forward layer. In contrast to conventional Transformers \cite{vaswani2017attention,sharma2018conceptual} that utilize mask to prevent positions from attending to subsequent positions, the multi-head self-attention layer in Diffusion Transformer is bi-directional without masks. In this way, the $i$-th Transformer decoder block operates as:
\begin{equation}
\begin{aligned}
&{{h^{i + 1}} = {\bf{FFN}}({\bf{norm}}({{\tilde h}^i} + {\bf{MultiHead}}({{\tilde h}^i},\mathcal{\tilde V},\mathcal{\tilde V})))},\\
&{{{\tilde h}^i} = {\bf{norm}}({h^i} + {\bf{MultiHead}}({h^i},{h^i},{h^i}))}.
\end{aligned}
\end{equation}
The inputs of the first Transformer decoder block at time step $t$ are the word tokens $h^0 = \{ w_0^t,w_1^t,...,w_{N_s}^t \}$. After stacking $N_t$ blocks, the hidden state outputted by the last block $h^{N_t}$ is utilized to predict the probability distribution of each output word, which is calculated by:
\begin{equation}
{p_i} = {\bf{softmax}}({W^T}h_i^{{N_t}}),
\end{equation}
where ${W^T}$ is the weight matrix, $h_i^{{N_t}}$ and $p_i \in \mathbb{R}^{\mathcal{W}} $ is the hidden state vector and probability distribution corresponding to the $i$-th word, respectively. After that, we map the probability distribution of $p_i$ into bits $b_i$ by taking weighted average over all the $\mathcal{W}$ bits in the vocabulary:
\begin{equation}
{b_i} = \sum\nolimits_{c = 1}^\mathcal{W} {p_i^c} {B^c},
\end{equation}
where $p_i^c$ is the $c$-th probability of $p_i$ and $B^c$ is the bit representation of the $c$-th word in the vocabulary. To speedup convergence of  Diffusion Transformer, we also integrate the diffusion process objective $\mathcal{L}_{bit}$ in Eq. (\ref{eq:diff}) with a typical cross entropy loss $\mathcal{L}_{XE}$ over the probability distribution $p$ during training. Therefore, the final objective is calculated as: $\mathcal{L} = \mathcal{L}_{XE} + \mathcal{L}_{bit}$.

\subsection{Semantic Condition}
\label{sec:diffusion_prior}
Recent practice attempts to employ diffusion model for image captioning by directly estimating $x_0$ from the latent state $x_T$ sampled from noise distribution $\mathcal{N}(0,\textit{I})$ in the reverse process. Nevertheless, this way overlooks the complex visual-language alignment and inherent sequential dependency among words during diffusion process, thereby resulting in the word repetition or omissions problems. To mitigate this limitation, we upgrade the diffusion model with additional semantic prior, yielding a semantic-conditioned diffusion process for Diffusion Transformer.

Technically, given an image, we first search the semantically relevant sentence from training sentence pool by using an off-the-shelf cross-modal retrieval model. The relevant sentence is further represented as a sequence of word tokens $s_r$, which reflect the comprehensive semantic information. After that, we take $s_r$ as semantic condition to constrain the diffusion process of Diffusion Transformer. In particular, at each time step $t$, we capitalize on an additional semantic Transformer with $N_p$ semantic Transformer blocks to contextually encode current latent state $x_t$ with semantic prior $s_r$. Here we concatenate $x_t$ with the previous timestep prediction ${{\tilde x}_0}$ along channel dimension as in \cite{analogbits}. The textual features of latent state $x_t$ and semantic prior $s_r$ are thus calculated as:
\begin{equation}
\label{eq:prior_input}
\begin{aligned}
&{z^x = {\bf{FC}}({\bf{Concat}}({x_t},{{\tilde x}_0})) + \varphi (\gamma (t'))},\\
&{{z^r} = {\bf{FC}}({s_r})}.
\end{aligned}
\end{equation}
where $\varphi$ is a multi-layer perception. The positional encodings are omitted for simplicity. Next, we concatenate the textual features $z^x$ and $z^r$ as (${\mathcal{W}^0}=[z^x,z^r]$), which is further fed into semantic Transformer to achieve the semantic-conditional latent state. In this way, the $i$-th semantic Transformer block is calculated as:
\begin{equation}
{\mathcal{W}^{i + 1}} = {\bf{FFN}}({\bf{norm}}({\mathcal{W}^i} + {\bf{MultiHead}}({\mathcal{W}^i},{\mathcal{W}^i},{\mathcal{W}^i})))
\end{equation}
Finally, by denoting the outputs of semantic Transformer as ${\mathcal{W}^{N_p}}=[{\mathcal{W}_{x}^{N_p}},{\mathcal{W}_{r}^{N_p}}]$, we take ${\mathcal{W}_{x}^{N_p}}$ as the strengthened semantic-conditional latent state ${\mathcal{W}_{x}^{N_p}}=h^0=\{ w_0^t,w_1^t,...,w_{N_s}^t \}$, which will be fed into sentence decoder for caption generation in diffusion model.


\subsection{Cascaded Diffusion Transformer}
Inspired by the success of cascaded diffusion models for image generation \cite{ho2022cascaded}, our SCD-Net stacks multiple Diffusion Transformers in a cascaded fashion. Such cascaded structure aims to progressively strengthen the output sentence with better visual-language alignment and linguistical coherence. Formally, this cascaded diffusion process can be represented as:
\begin{equation}
F(x_t, \gamma (t'), \mathcal{V}) = {f_M} \circ {f_{M - 1}} \circ  \cdot  \cdot  \cdot  \circ {f_1}(x_t, \gamma (t'), \mathcal{V}),
\end{equation}
where $M$ is the total number of stacked Diffusion Transformer, and ${f_1}$ is the first Diffusion Transformer equipped with aforementioned semantic condition. In this way, each Diffusion Transformer ${f_i}$ ($i \ge 2$) operates diffusion process conditioned on the sentence prediction $x^{i-1}_0$ of previous Diffusion Transformer $f_{i-1}$. Accordingly, for each Diffusion Transformer ${f_i}$ ($i \ge 2$), we slightly modify its structure to take the additional semantic cues of $x^{i-1}_0$ into account.
Specifically, given the latent state $x_t$, previous timestep prediction ${{\tilde x}_0}$, and the previous Diffusion Transformer prediction $x^{i-1}_0$, we remould Eq. (\ref{eq:prior_input}) by measuring textual features of latent state $z^x$ as:
\begin{equation}
z^x = {\bf{FC}}({\bf{Concat}}({x_t},{{\tilde x}_0}, x^{i-1}_0)) + \varphi (\gamma (t')).
\end{equation}
Then the textual features $z^x$ are concatenated with semantic prior $z^r$, and we feed them into the semantic Transformer. At inference, the output of each Diffusion Transformer ${f_i}$ is directly fused with the output of previous Diffusion Transformer $f_{i-1}$ at each timestep.

\subsection{Guided Self-Critical Sequence Training}
Conventional autoregressive image captioning techniques \cite{anderson2017bottom,sharma2018conceptual,rennie2017self} commonly utilize the Self-Critical Sequence Training \cite{rennie2017self} (SCST) to boost up performances with sentence-level optimization (e.g., CIDEr metric \cite{vedantam2015cider}):
\begin{equation}
{\mathcal{L}_R}(\theta ) =  - {{\rm E}_{{y_{1:{N_s}}} \sim {p_\theta }}}[R({y_{1:{N_s}}})],
\end{equation}
where $R$ denotes the CIDEr score function. The gradient of the loss can be approximated as:
\begin{equation}
{\nabla _\theta }{\mathcal{L}_R}(\theta ) \approx  - (R(y_{1:N_s}^s) - R({\hat y_{1:N_s}})){\nabla _\theta }\log {p_\theta }(y_{1:N_s}^s),
\end{equation}
where $y_{1:N_s}^s$ is the sampled caption and $R({\hat y_{1:N_s}})$ denotes the sentence-level reward of baseline in greedily decoding inference.
However, it is not trivial to directly apply SCST to the diffusion process in Diffusion Transformer. The difficulty mainly originates from two aspects. First, the non-autoregressive inference procedure of Diffusion Transformer contains multiple steps (e.g., more than 20), and thus it is impractical to sample sentences directly from the noise $x_T$. Furthermore, since each output word in the Diffusion Transformer is sampled independently, simply assigning the same reward to each word would result in typical word repetition problem \cite{guo2021non}. To address these limitations, we propose a new Guided Self-Critical Sequence Training, which nicely guides the learning of SCD-Net with the knowledge derived from a standard autoregressive Transformer model.


Technically, we first train a standard autoregressive Transformer teacher model that shares same architecture of Diffusion Transformer. Next, for each training image, this Transformer teacher model predicts high-quality sentence $S^{tea}$ as additional semantic guidance. Then we feed the predicted high-quality sentence $S^{tea}$ into the cascaded Diffusion Transformer, instead of the ground-truth sentences. Rather than randomly sampling several captions as in conventional SCST, we enforce one of the sampled sentences to be the same as the predicted sentence $S^{tea}$. Let ${y'}_{1:N_s}^{s_j}|_{j=0}^{N_y}$ denote the sampled sentences containing $S^{tea}$, where $N_y$ is the number of sampled sentences. The gradient of ${\mathcal{L}_R}(\theta )$ is thus measured as:
\begin{equation}\small
{\nabla _\theta }{\mathcal{L}_R}(\theta ) \approx  - \frac{1}{{{N_y}}}\sum\limits_{j = 0}^{{N_y}} {(R({y'}_{1:{N_s}}^{{s_j}}) - R({{\hat y}_{1:{N_s}}})){\nabla _\theta }\log {p_\theta }({y'}_{1:{N_s}}^{{s_j}})}.
\end{equation}
In this way, the sampling of the high-quality sentence $S^{tea}$ tends to receive a positive reward, that encourages Diffusion Transformer to produce high-quality sentence. The possibility of other low-quality sentences (e.g., sentences with word repetition) is thus suppressed. It is worthy to note that when the diffusion model becomes saturated, we replace the sentence $S^{tea}$ derived from autoregressive Transformer model with the estimated $S'$ by the diffusion model, if the quality of $S'$ (measured in CIDEr) is higher than $S^{tea}$.

\section{Experiments}
In this section, we empirically verify the effectiveness of our SCD-Net by conducting experiments over the widely adopted COCO benchmark \cite{Lin:ECCV14} for image captioning.

\begin{table*}[t]
  \centering
  \vspace{-0.00in}
  \setlength\tabcolsep{3.0pt}
  \caption{Comparison results of SCD-Net with other state-of-the-art autoregressive and non-autoregressive approaches on COCO Karpathy test split for image captioning. $\dag$ denotes the use of a superior object detector Pix2seq \cite{chen2021pix2seq}.}
  \vspace{-0.0in}
  \begin{tabular}{l|cccccccc|cccccccc}
  \Xhline{2\arrayrulewidth}
               & \multicolumn{8}{c|}{\textbf{Cross-Entropy Loss}} & \multicolumn{8}{c}{\textbf{CIDEr Score Optimization}} \\
               & B@1  & B@2  & B@3 & B@4 & M & R & C & S & B@1  & B@2  & B@3  & B@4  & M  & R  & C  & S \\ \hline
\multicolumn{17}{c}{\textbf{Autoregressive}} \\ \hline
SCST \cite{rennie2017self}        &   -  &   -  &   -  & 30.0 & 25.9 & 53.4 & 99.4  &  -   &  -   &  -   &  -   & 34.2 & 26.7 & 55.7 & 114.0 &  -  \\
RFNet \cite{jiang2018recurrent}   & 76.4 & 60.4 & 46.6 & 35.8 & 27.4 & 56.5 & 112.5 & 20.5 & 79.1 & 63.1 & 48.4 & 36.5 & 27.7 & 57.3 & 121.9 & 21.2 \\
Up-Down \cite{anderson2017bottom} & 77.2 &   -  &   -  & 36.2 & 27.0 & 56.4 & 113.5 & 20.3 & 79.8 &  -   &  -   & 36.3 & 27.7 & 56.9 & 120.1 & 21.4 \\
GCN-LSTM \cite{yao2018exploring}  & 77.3 &   -  &   -  & 36.8 & 27.9 & 57.0 & 116.3 & 20.9 & 80.5 &  -   &  -   & 38.2 & 28.5 & 58.3 & 127.6 & 22.0 \\
LBPF \cite{qin2019look}  & 77.8 &   -  &   -  & 37.4 & 28.1 & 57.5 & 116.4 & 21.2 & 80.5 &  -   &  -   & 38.3 & 28.5 & 58.4 & 127.6 & 22.0 \\
SGAE \cite{Yang:CVPR19}  & 77.6 &   -  &   -  & 36.9 & 27.7 & 57.2 & 116.7 & 20.9 & 80.8 &  -   &  -   & 38.4 & 28.4 & 58.6 & 127.8 & 22.1 \\
ORT \cite{herdade2019image}       &  -   &   -  &   -  &   -  &   -  &   -  &    -  &  -   & 80.5 &  -   &  -   & 38.6 & 28.7 & 58.4 & 128.3 & 22.6 \\
Transformer \cite{sharma2018conceptual} & 76.1 & 59.9 & 45.2 & 34.0 & 27.6 & 56.2 & 113.3 & 21.0 & 80.2 & 64.8 & 50.5 & 38.6 & 28.8 & 58.5 & 128.3 & 22.6 \\
AoANet \cite{huang2019attentio}   & 77.4 &   -  &   -  & 37.2 & 28.4 & 57.5 & 119.8 & 21.3 & 80.2 &  -   &  -   & 38.9 & 29.2 & 58.8 & 129.8 & 22.4 \\
$M^2$ Transformer \cite{cornia2020meshed} &  - & -  & -  & -  &   -  &  -   &   -   &   -  & 80.8 &  -   &  -   & 39.1 & 29.2 & 58.6 & 131.2 & 22.6 \\ \hline
\multicolumn{17}{c}{\textbf{Non-Autoregressive}} \\ \hline
MNIC \cite{gao2019masked}      & 75.4 & 57.7 & 42.6 & 30.9 & 27.5 & 55.6 & 108.1 & 21.0 &  -   & -    & -    & -    &   -  &  -   &   -   & -   \\
MIR \cite{lee2020det}          & -    &  -   & -    & 32.5 & 27.2 &  -   & 109.5 & 20.6 &  -   & -    & -    & -    &   -  &  -   &   -   & -   \\
CMAL \cite{guo2021non}         & 78.5 &  -   & -    & 35.3 & 27.3 & 56.9 & 115.5 & 20.8 & 80.3 & -    & -    & 37.3 & 28.1 & 58.0 & 124.0 & 21.8 \\
SATIC \cite{zhou2021semi}       & 77.3 &  -   & -    & 32.9 & 27.0 & -    & 111.0 & 20.5 & 80.6 & -    & -    & 37.9 & 28.6 & -    & 127.2 & 22.3 \\
Bit Diffusion \textdagger \cite{analogbits} & -    & -    & -    & 34.7 & -    & \textbf{58.0} & 115.0 & -   &  -   & -    & -    & -    &   -  &  -   &   -   & -   \\
SCD-Net                        & \textbf{79.0} & \textbf{63.4} & \textbf{49.1} & \textbf{37.3} & \textbf{28.1} & \textbf{58.0} & \textbf{118.0} & \textbf{21.6} & \textbf{81.3} & \textbf{66.1} & \textbf{51.5} & \textbf{39.4} & \textbf{29.2} & \textbf{59.1} & \textbf{131.6} & \textbf{23.0} \\
  \Xhline{2\arrayrulewidth}
  \end{tabular}
  \vspace{-0.1in}
  \label{tab:COCO_single}
\end{table*}

\subsection{Datasets and Experimental Settings}
\textbf{Dataset.}
COCO is the most popular dataset in image captioning field, which comprises 82,783 training images, 40,504 validation images, and 40,775 testing images. Each image is annotated with five descriptions. Considering that the annotations of the official testing set are not provided, we utilize the widely adopted Karpathy split \cite{rennie2017self,anderson2017bottom} and take the 113,287 images for training, 5,000 images for validation, and 5,000 image for testing. Moreover, we also report the performances on the official testing set through online testing server.
We convert all the training sentences into lower case and filter out rare words which occur less than four times following \cite{anderson2017bottom}. In this way, the final vocabulary consists of 10,199 unique words.

\textbf{Implementation Details.} The whole SCD-Net is implemented in X-modaler codebase \cite{li2021x}. In SCD-Net, we employ the Faster-RCNN pre-trained on ImageNet \cite{imagenet} and Visual Genome \cite{krishna2017visual} to extract image region features. The dimension of the original region feature is 2,048 and we transform it into a 512-dimensional feature by a fully-connected layer. Each word is converted into 14-bits as in \cite{analogbits}. The visual encoder, sentence decoder, and the semantic Transformer contains $N_v=3$, $N_t=3$, and $N_p=3$ Transformer blocks. The hidden size of each Transformer block is set as 512. The training of SCD-Net includes two stages. In the first stage, the whole architecture of SCD-Net is optimized via Adam \cite{kingma2014adam} optimizer on four V100 GPUs with $\ell _2$ regression loss and labels smoothing. The optimization process includes 60 epochs with a batch size of 16 and the same learning rate scheduling strategy as in \cite{vaswani2017attention} (warmup: 20,000 iterations). In the second stage, we select the initialization model trained by the first stage that achieves the best CIDEr score on validation set. Then SCD-Net is optimized with CIDEr score via our guided self-critical sequence training for 60 epochs. We fix the batch size and learning rate as 16 and 0.00001, respectively. At inference, the time step number and the time difference is set as 50 and 0, respectively. Five types of evaluation metrics are adopted: BLEU@N \cite{Papineni:ACL02} (B@1-4), METEOR \cite{Banerjee:ACL05} (M), ROUGE \cite{lin2004rouge} (R), CIDEr \cite{vedantam2015cider} (C), and SPICE \cite{spice2016} (S). All the metrics are measured through the source codes released by COCO Evaluation Server \cite{chen2015microsoft}.

\subsection{Performance Comparison}
\textbf{Offline Evaluation.}
Table \ref{tab:COCO_single} summaries the performances of our SCD-Net on the offline COCO Karpathy test split under two different optimization strategies, i.e., the standard cross-entropy loss and the sentence-level optimization in CIDEr score.
Considering that our SCD-Net belongs to non-autoregressive image captioning techniques, here we include two groups of baselines (autoregressive and non-autoregressive methods).

\begin{table*}[!tb]
  \setlength\tabcolsep{5.5pt}
  \centering
  \caption{Comparison results of SCD-Net with other state-of-the-art approaches on the official test split in online test server.}
  \label{table:leaderboard}
  \vspace{-0.00in}
  \begin{tabular}{l|*{13}{c|}c}
  \Xhline{2\arrayrulewidth}
      \multicolumn{1}{c|}{\multirow{2}{*}{{Model}}} & \multicolumn{2}{c|}{{B@1}} & \multicolumn{2}{c|}{{B@2}} & \multicolumn{2}{c|}{{B@3}} & \multicolumn{2}{c|}{{B@4}} & \multicolumn{2}{c|}{{M}} & \multicolumn{2}{c|}{{R}} & \multicolumn{2}{c}{{C}} \\\cline{2-15}
      \multicolumn{1}{c|}{}&c5 &c40 &c5 &c40 &c5 &c40&c5 &c40&c5 &c40&c5 &c40&c5 &c40 \\\hline
  \multicolumn{15}{c}{\textbf{Ensemble Model}} \\ \hline
  {SCST} \cite{rennie2017self}            & 78.1 & 93.7 & 61.9 & 86.0 & 47.0 & 75.9 & 35.2 & 64.5 & 27.0 & 35.5 & 56.3 & 70.7 & 114.7 & 116.7 \\
  {Up-Down} \cite{anderson2017bottom}     & 80.2 & 95.2 & 64.1 & 88.8 & 49.1 & 79.4 & 36.9 & 68.5 & 27.6 & 36.7 & 57.1 & 72.4 & 117.9 & 120.5 \\
  {RFNet} \cite{jiang2018recurrent}       & 80.4 & 95.0 & 64.9 & 89.3 & 50.1 & 80.1 & 38.0 & 69.2 & 28.2 & 37.2 & 58.2 & 73.1 & 122.9 & 125.1 \\
  {GCN-LSTM} \cite{yao2018exploring}          & 80.8 & 95.2 & 65.5 & 89.3 & 50.8 & 80.3 & 38.7 & 69.7 & 28.5 & 37.6 & 58.5 & 73.4 & 125.3 & 126.5 \\
  APN \cite{yang2021auto}                     &  -   &  -   &  -   &  -   &  -   &  -   & 38.9 & 70.2 & 28.8 & 38.0 & 58.7 & 73.7 & 126.3 & 127.6 \\
  {AoANet} \cite{huang2019attentio}           & 81.0 & 95.0 & 65.8 & 89.6 & 51.4 & 81.3 & 39.4 & 71.2 & 29.1 & 38.5 & 58.9 & 74.5 & 126.9 & 129.6 \\ \hline
  \multicolumn{15}{c}{\textbf{Single Model}} \\ \hline
  CMAL \cite{guo2021non}                    & 79.8 & 94.3 & 63.8 & 87.2 & 48.8 & 77.2 & 36.8 & 66.1 & 27.9 & 36.4 & 57.6 & 72.0 & 119.3 & 121.2 \\
  CAVP \cite{liu2018context}                & 80.1 & 94.9 & 64.7 & 88.8 & 50.0 & 79.7 & 37.9 & 69.0 & 28.1 & 37.0 & 58.2 & 73.1 & 121.6 & 123.8 \\
  {SGAE} \cite{Yang:CVPR19}                 & \textbf{80.6} & 95.0 & \textbf{65.0} & 88.9 & \textbf{50.1} & 79.6 & 37.8 & 68.7 & 28.1 & 37.0 & 58.2 & 73.1 & 122.7 & 125.5 \\
  SCD-Net                                   & 80.2 & \textbf{95.1} & 64.9 & \textbf{89.3} & \textbf{50.1} & \textbf{80.1} & \textbf{38.1} & \textbf{69.4} & \textbf{29.0} & \textbf{38.2} & \textbf{58.5} & \textbf{73.5} & \textbf{126.2} & \textbf{129.2} \\
  \Xhline{2\arrayrulewidth}
  \end{tabular}
  \vspace{-0.18in}
\end{table*}

In general, there is a clear performance gap between the most state-of-the-art autoregressive and non-autoregressive baselines under both optimization strategies. The results basically verify the weakness of non-autoregressive techniques that incurs word repetition or omissions problems. In between, even a basic autoregressive Transformer encoder-decoder model (Transformer \cite{sharma2018conceptual}) can achieve comparable performances with the best competitors in non-autoregressive baselines (Bit Diffusion and SATIC). In contrast, by constraining the typical diffusion model with additional semantic prior, our SCD-Net consistently exhibits better performances than autoregressive Transformer across all metrics. In particular, the CIDEr and SPICE score of SCD-Net under cross-entropy loss is 118.0\% and 21.6\%, which manifest the absolute improvement of 4.7\% and 0.6\% against autoregressive Transformer, respectively. This demonstrates the key advantage of our semantic-conditional diffusion model for image captioning in a non-autoregressive fashion. Nevertheless, the performances of our SCD-Net under cross-entropy loss are still inferior to state-of-the-art autoregressive methods with deliberate designs of upgraded Transformer structures (e.g., attention on attention in AoANet and meshed memory in $M^2$ Transformer).
When further equipped with guided self-critical sequence training, our SCD-Net manages to outperform the most competitive autoregressive baseline with an upgraded Transformer structure ($M^2$ Transformer), leading to slight improvements in CIDEr (0.4\%) and SPICE (0.4\%). Such encouraging performances confirm the effectiveness of guided self-critical sequence training tailored for diffusion model in SCD-Net.

\textbf{Online Evaluation.}
In addition, we also evaluate our SCD-Net on the official testing set by submitting the results to online testing server\footnote{\url{https://competitions.codalab.org/competitions/3221}}.
Table \ref{table:leaderboard} shows the performance comparisons. It is worthy to note that here we only report the performances of our SCD-Net with single model, without using any model ensemble as in some baselines (e.g., APN and AoANet).
Similar to the observations in offline evaluation, our SCD-Net again surpasses all the single-model baselines across most metrics, including both autoregressive methods (e.g., SGAE) and non-autoregressive approach (CMAL). Moreover, the single model of our SCD-Net even attains comparable performances with the ensemble version of some state-of-the-art autoregressive methods (e.g., APN). The results basically prove the advantage of exploiting semantic condition in diffusion model for image~captioning.

\begin{figure}[!tb]
 \vspace{-0.0in}
    \centering {\includegraphics[width=0.49\textwidth]{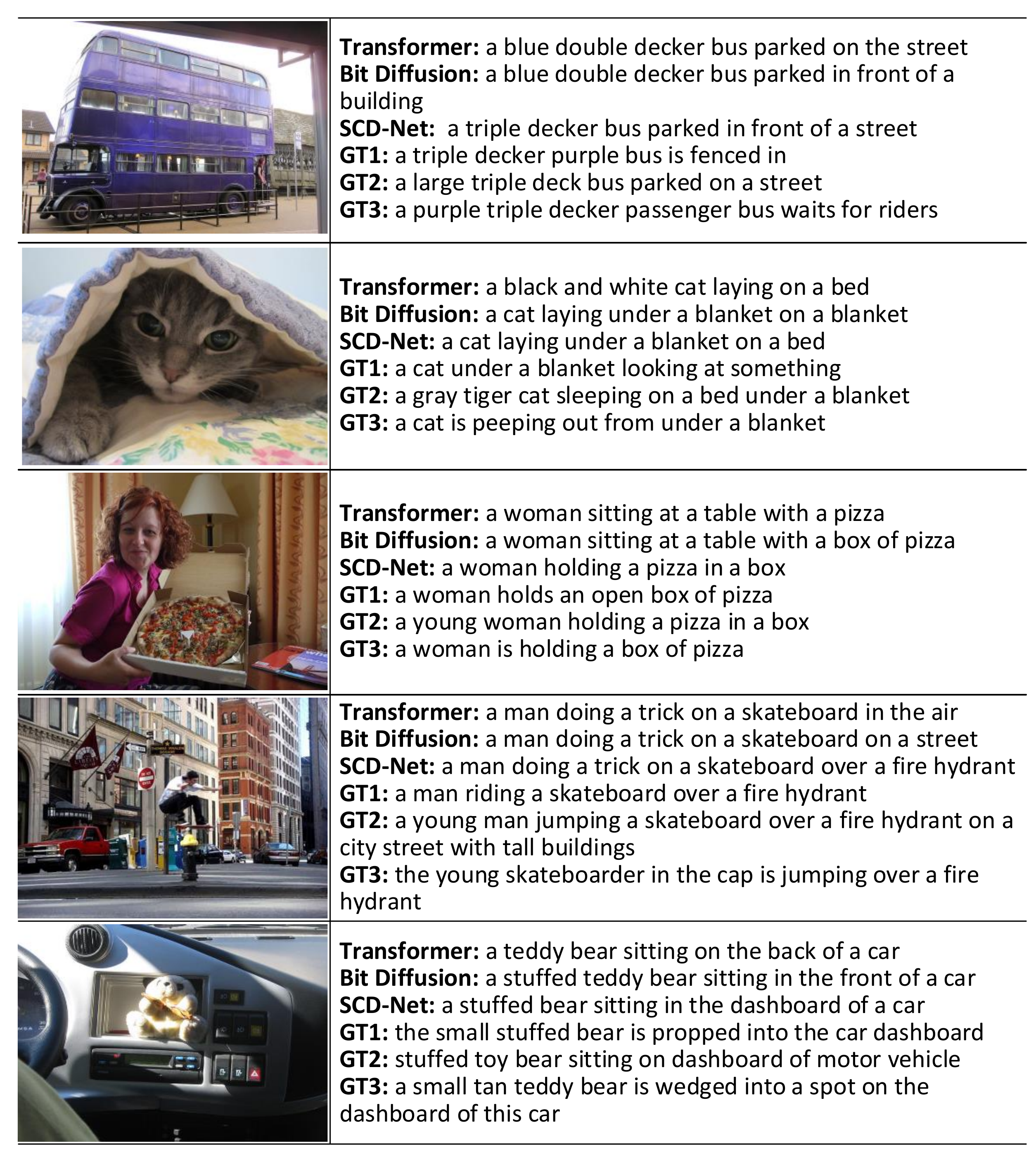}}
    \vspace{-0.3in}
    \caption{Examples of image captioning results generated by Transformer \cite{sharma2018conceptual}, Bit Diffusion \cite{analogbits} and our SCD-Net, as well as the corresponding ground-truths (GT).}
    \label{fig:case}
    \vspace{-0.1in}
\end{figure}

\textbf{Qualitative Analysis.}
Figure \ref{fig:case} further illustrates the image captioning ground-truths and qualitative results of different approaches for five different images. In general,
all the three methods are able to produce somewhat semantically relevant and linguistically coherent captions. However, the explicit modeling of key semantics in an image is overlooked in Transformer, which might result in word omissions (e.g., the missing of ``blanket'' in the first image). Moreover, the non-autoregressive Bit Diffusion sometimes produces low-quality caption with word repetition problem (e.g., the repetition of ``blanket'' in the first image). To address these issues, our SCD-Net novelly capitalizes on semantic-conditioned diffusion model and guided self-critical sequence training to facilitate image captioning, leading to more accurate and descriptive captions.

\begin{table}[t]
  \centering
  \vspace{-0.0in}
  \setlength\tabcolsep{1.2pt}
  \caption{Ablation study on each design in SCD-Net on COCO Karpathy test split. \textbf{Base} denotes the conventional diffusion model. \textbf{Semantic} represents the use of semantic condition in diffusion model. \textbf{RL} refers to the sentence-level optimization in reinforcement learning, which can be set as the typical Self-Critical Sequence Training (SCST) \cite{rennie2017self} or our proposed Guided SCST (GSCST). \textbf{Cascade} denotes the use of the cascaded structure.}
    \label{table:ablation}
  \vspace{-0.0in}
  \begin{tabular}{cccc|ccccc}
  \Xhline{2\arrayrulewidth}
  Base         & Semantic     &   RL           & Cascade           & B@4  & M    & R    & C     & S \\ \hline
  $\checkmark$ &              &                &                   & 35.9 & 27.3 & 57.2 & 114.5 & 20.7 \\
  $\checkmark$ & $\checkmark$ &                &                   & 36.4 & 27.8 & 57.4 & 116.2 & 21.2 \\
  $\checkmark$ & $\checkmark$ &   SCST         &                   & 34.6 & 27.7 & 57.1 & 120.8 & 21.5 \\
  $\checkmark$ & $\checkmark$ &  GSCST         &                   & 38.5 & 29.1 & 58.6 & 128.6 & 22.9 \\
  $\checkmark$ & $\checkmark$ &  GSCST         &  $\checkmark$     & 39.4 & 29.2 & 59.1 & 131.6 & 23.0 \\ \Xhline{2\arrayrulewidth}
  \end{tabular}
    \vspace{-0.2in}
\end{table}

\subsection{Experimental Analysis}

\textbf{Ablation Study.} Here we investigate how each design in our SCD-Net influences the overall image captioning performance on COCO Karpathy test split. Table \ref{table:ablation} details the performances across different ablated runs of SCD-Net. We start from a base model (\textbf{Base}) that directly leverages the typical diffusion process as in \cite{analogbits} to train single Diffusion Transformer, which achieves 114.5\% in CIDEr under cross-entropy loss. Note that our implemented Base model is slightly inferior to primary Bit Diffusion (CIDEr: 115.0\%) which is equipped with a stronger object detector. Next, by upgrading conventional diffusion model with the additional semantic prior (\textbf{Semantic}), we observe clear performance boosts for such semantic-conditional diffusion model. When further training our Diffusion Transformer with sentence-level optimization (\textbf{RL}) via \textbf{SCST}, only the CIDEr and SPICE scores are slightly improved, while the other scores even decrease. The results basically reveal the difficulty of applying SCST to non-autoregressive solution, due to the indisposed
sentence quality of randomly sampled captions in SCST. As an alternative, our Guided SCST (\textbf{GSCST}) guides the reinforcement learning of Diffusion Transformer with high-quality sentence derived from an autoregressive Transformer teacher model, thereby leading to significant improvements in terms of all metrics. Finally, after stacking multiple Diffusion Transformers in a cascaded manner (\textbf{Cascade}), our SCD-Net obtains the best image captioning performance.

\textbf{Ablation on Different Transformer Block Number.} We further examine how image captioning performances are affected when capitalizing on different numbers of Transformer blocks in single Diffusion Transformer. As shown in Table \ref{table:block}, the performances are relatively smooth when the Transformer block number varies between 3 and 4. When enlarging the block number to 5, the performances slightly decreases. We speculate that this may be the results of unnecessary context information mining among the input tokens with more Transformer blocks in visual encoder and sentence decoder. Accordingly, the Transformer block number in Diffusion Transformer is set as 3, which achieves the best performances with less computational cost.

\begin{table}[t]
  \centering
  \vspace{-0.0in}
    \setlength\tabcolsep{2.6pt}
  \caption{Ablation on different Transformer block number in single Diffusion Transformer (without cascaded structure) on COCO Karpathy test split.}
  \vspace{-0.0in}
  \begin{tabular}{c|ccccc}
  \Xhline{2\arrayrulewidth}
  \# Transformer Block  & B@4   & M    & R     & C      & S    \\ \hline
     3      & 38.5  & 29.1 & 58.6  & 128.6  & 22.9 \\
     4      & 38.3  & 29.1 & 58.5  & 128.6  & 23.0 \\
     5      & 38.4  & 29.0 & 58.5  & 128.1  & 22.9 \\
     6      & 38.2  & 29.0 & 58.4  & 128.1  & 22.9 \\  \Xhline{2\arrayrulewidth}
  \end{tabular}
       \label{table:block}
\end{table}

\begin{table}[t]
  \centering
  \vspace{-0.0in}
  \setlength\tabcolsep{2.6pt}
  \caption{Ablation on different Diffusion Transformer number in SCD-Net (with cascaded structure) on COCO Karpathy test split.}
  \vspace{-0.0in}
  \begin{tabular}{c|ccccc}
  \Xhline{2\arrayrulewidth}
  \# Diffusion Transformer   & B@4   & M     & R     & C       & S    \\ \hline
     1      & 38.5  & 29.1  & 58.6  & 128.6   & 22.9 \\
     2      & 39.4  & 29.2  & 59.1  & 131.6   & 23.0 \\
     3      & 39.5  & 29.2  & 59.1  & 131.7   & 23.0 \\ \Xhline{2\arrayrulewidth}
  \end{tabular}
    \label{table:transformer}
        \vspace{-0.2in}
\end{table}

\textbf{Ablation on Different Diffusion Transformer Number in SCD-Net.} To explore the relationship between the performance and the Diffusion Transformer number in SCD-Net, we show the performances by varying this number from 1 to 3. As listed in Table \ref{table:transformer}, more stacked Diffusion Transformer in a cascaded manner generally leads to performance improvements. This finding basically highlights the merit of progressively strengthening the output sentence in a cascaded fashion. In particular, once the number of Diffusion Transformer is larger than 1, the performances are less affected, easing the selection of Diffusion Transformer number in SCD-Net practically. Finally, we empirically set the Diffusion Transformer number as 2.

\section{Conclusion}

In this work, we delve into the idea of strengthening visual-language alignment and linguistical coherence in diffusion model for image captioning.
To verify our claim, we shape a new semantic-conditional diffusion process that upgrades diffusion model with additional semantic prior. A guided self-critical sequence training strategy is further devised to stabilize and boost the diffusion process. We empirically validate the superiority of our proposal against state-of-the-art non-autoregressive approaches. At the same time, we are happy to see that our new diffusion model based paradigm manages to outperform the competitive autoregressive method sharing the same Transformer encoder-decoder structure. The results basically demonstrate the encouraging potential of diffusion model in image captioning.

{\small
\bibliographystyle{ieee_fullname}
\bibliography{egbib}

\begin{thebibliography}{10}\itemsep=-1pt

\bibitem{spice2016}
Peter Anderson, Basura Fernando, Mark Johnson, and Stephen Gould.
\newblock Spice: Semantic propositional image caption evaluation.
\newblock In {\em ECCV}, 2016.

\bibitem{anderson2017bottom}
Peter Anderson, Xiaodong He, Chris Buehler, Damien Teney, Mark Johnson, Stephen
  Gould, and Lei Zhang.
\newblock Bottom-up and top-down attention for image captioning and visual
  question answering.
\newblock In {\em CVPR}, 2018.

\bibitem{Bahdanau14}
Dzmitry Bahdanau, Kyunghyun Cho, and Yoshua Bengio.
\newblock Neural machine translation by jointly learning to align and
  translate.
\newblock In {\em ICLR}, 2015.

\bibitem{Banerjee:ACL05}
Satanjeev Banerjee and Alon Lavie.
\newblock Meteor: An automatic metric for mt evaluation with improved
  correlation with human judgments.
\newblock In {\em ACL workshop}, 2005.

\bibitem{chen2021pix2seq}
Ting Chen, Saurabh Saxena, Lala Li, David~J Fleet, and Geoffrey Hinton.
\newblock Pix2seq: A language modeling framework for object detection.
\newblock In {\em ICLR}, 2021.

\bibitem{analogbits}
Ting Chen, Ruixiang Zhang, and Geoffrey Hinton.
\newblock Analog bits: Generating discrete data using diffusion models with
  self-conditioning.
\newblock {\em arXiv preprint arXiv:2208.04202}, 2022.

\bibitem{chen2015microsoft}
Xinlei Chen, Hao Fang, Tsung-Yi Lin, Ramakrishna Vedantam, Saurabh Gupta, Piotr
  Doll{\'a}r, and C~Lawrence Zitnick.
\newblock Microsoft coco captions: Data collection and evaluation server.
\newblock {\em arXiv preprint arXiv:1504.00325}, 2015.

\bibitem{cornia2020meshed}
Marcella Cornia, Matteo Stefanini, Lorenzo Baraldi, and Rita Cucchiara.
\newblock Meshed-memory transformer for image captioning.
\newblock In {\em CVPR}, 2020.

\bibitem{imagenet}
Jia Deng, Wei Dong, Richard Socher, Li-Jia Li, Kai Li, and Li Fei-Fei.
\newblock Imagenet: A large-scale hierarchical image database.
\newblock In {\em CVPR}, 2009.

\bibitem{donahue2015long}
Jeffrey Donahue, Lisa Anne~Hendricks, Sergio Guadarrama, Marcus Rohrbach,
  Subhashini Venugopalan, Kate Saenko, and Trevor Darrell.
\newblock Long-term recurrent convolutional networks for visual recognition and
  description.
\newblock In {\em CVPR}, 2015.

\bibitem{gao2019masked}
Junlong Gao, Xi Meng, Shiqi Wang, Xia Li, Shanshe Wang, Siwei Ma, and Wen Gao.
\newblock Masked non-autoregressive image captioning.
\newblock {\em arXiv preprint arXiv:1906.00717}, 2019.

\bibitem{nat2018}
Jiatao Gu, James Bradbury, Caiming Xiong, Victor~OK Li, and Richard Socher.
\newblock Non-autoregressive neural machine translation.
\newblock In {\em ICLR}, 2018.

\bibitem{guo2021non}
Longteng Guo, Jing Liu, Xinxin Zhu, Xingjian He, Jie Jiang, and Hanqing Lu.
\newblock Non-autoregressive image captioning with counterfactuals-critical
  multi-agent learning.
\newblock In {\em IJCAI}, 2021.

\bibitem{herdade2019image}
Simao Herdade, Armin Kappeler, Kofi Boakye, and Joao Soares.
\newblock Image captioning: Transforming objects into words.
\newblock {\em NeurIPS}, 2019.

\bibitem{ddpm}
Jonathan Ho, Ajay Jain, and Pieter Abbeel.
\newblock Denoising diffusion probabilistic models.
\newblock In {\em NeurIPS}, 2020.

\bibitem{ho2022cascaded}
Jonathan Ho, Chitwan Saharia, William Chan, David~J Fleet, Mohammad Norouzi,
  and Tim Salimans.
\newblock Cascaded diffusion models for high fidelity image generation.
\newblock {\em Journal of Machine Learning Research}, 2022.

\bibitem{huang2019attentio}
Lun Huang, Wenmin Wang, Jie Chen, and Xiao-Yong Wei.
\newblock Attention on attention for image captioning.
\newblock In {\em ICCV}, 2019.

\bibitem{jiang2018recurrent}
Wenhao Jiang, Lin Ma, Yu-Gang Jiang, Wei Liu, and Tong Zhang.
\newblock Recurrent fusion network for image captioning.
\newblock In {\em ECCV}, 2018.

\bibitem{karpathy2015deep}
Andrej Karpathy and Li Fei-Fei.
\newblock Deep visual-semantic alignments for generating image descriptions.
\newblock In {\em CVPR}, 2015.

\bibitem{kingma2014adam}
Diederik Kingma and Jimmy Ba.
\newblock Adam: A method for stochastic optimization.
\newblock In {\em ICLR}, 2015.

\bibitem{vdm}
Diederik Kingma, Tim Salimans, Ben Poole, and Jonathan Ho.
\newblock Variational diffusion models.
\newblock In {\em NeurIPS}, 2021.

\bibitem{krishna2017visual}
Ranjay Krishna, Yuke Zhu, Oliver Groth, Justin Johnson, Kenji Hata, Joshua
  Kravitz, Stephanie Chen, Yannis Kalantidis, Li-Jia Li, David~A Shamma, et~al.
\newblock Visual genome: Connecting language and vision using crowdsourced
  dense image annotations.
\newblock {\em IJCV}, 2017.

\bibitem{lee2020det}
Jason Lee, Elman Mansimov, and Kyunghyun Cho.
\newblock Deterministic non-autoregressive neural sequence modeling by
  iterative refinement.
\newblock In {\em EMNLP}, 2020.

\bibitem{li2021x}
Yehao Li, Yingwei Pan, Jingwen Chen, Ting Yao, and Tao Mei.
\newblock X-modaler: A versatile and high-performance codebase for cross-modal
  analytics.
\newblock In {\em ACM MM}, 2021.

\bibitem{li2022comprehending}
Yehao Li, Yingwei Pan, Ting Yao, and Tao Mei.
\newblock Comprehending and ordering semantics for image captioning.
\newblock In {\em CVPR}, 2022.

\bibitem{li2019pointing}
Yehao Li, Ting Yao, Yingwei Pan, Hongyang Chao, and Tao Mei.
\newblock Pointing novel objects in image captioning.
\newblock In {\em CVPR}, 2019.

\bibitem{lin2004rouge}
Chin-Yew Lin.
\newblock Rouge: A package for automatic evaluation of summaries.
\newblock In {\em ACL Workshop}, 2004.

\bibitem{Lin:ECCV14}
Tsung-Yi Lin, Michael Maire, Serge Belongie, James Hays, Pietro Perona, Deva
  Ramanan, Piotr Doll{\'a}r, and C~Lawrence Zitnick.
\newblock Microsoft coco: Common objects in context.
\newblock In {\em ECCV}, 2014.

\bibitem{liu2018context}
Daqing Liu, Zheng-Jun Zha, Hanwang Zhang, Yongdong Zhang, and Feng Wu.
\newblock Context-aware visual policy network for sequence-level image
  captioning.
\newblock In {\em Proceedings of the 26th ACM international conference on
  Multimedia}, pages 1416--1424, 2018.

\bibitem{liu2021o2na}
Fenglin Liu, Xuancheng Ren, Xian Wu, Bang Yang, Shen Ge, and Xu Sun.
\newblock O2na: An object-oriented non-autoregressive approach for controllable
  video captioning.
\newblock In {\em Findings of the Association for Computational Linguistics:
  ACL-IJCNLP 2021}, 2021.

\bibitem{pan2020x}
Yingwei Pan, Ting Yao, Yehao Li, and Tao Mei.
\newblock X-linear attention networks for image captioning.
\newblock In {\em CVPR}, 2020.

\bibitem{Papineni:ACL02}
Kishore Papineni, Salim Roukos, Todd Ward, and Wei-Jing Zhu.
\newblock Bleu: a method for automatic evaluation of machine translation.
\newblock In {\em ACL}, 2002.

\bibitem{qin2019look}
Yu Qin, Jiajun Du, Yonghua Zhang, and Hongtao Lu.
\newblock Look back and predict forward in image captioning.
\newblock In {\em CVPR}, 2019.

\bibitem{ren2015faster}
Shaoqing Ren, Kaiming He, Ross Girshick, and Jian Sun.
\newblock Faster r-cnn: Towards real-time object detection with region proposal
  networks.
\newblock In {\em NeurIPS}, 2015.

\bibitem{rennie2017self}
Steven~J Rennie, Etienne Marcheret, Youssef Mroueh, Jerret Ross, and Vaibhava
  Goel.
\newblock Self-critical sequence training for image captioning.
\newblock In {\em CVPR}, 2017.

\bibitem{sharma2018conceptual}
Piyush Sharma, Nan Ding, Sebastian Goodman, and Radu Soricut.
\newblock Conceptual captions: A cleaned, hypernymed, image alt-text dataset
  for automatic image captioning.
\newblock In {\em ACL}, 2018.

\bibitem{Sutskever:NIPS14}
Ilya Sutskever, Oriol Vinyals, and Quoc~V Le.
\newblock Sequence to sequence learning with neural networks.
\newblock In {\em NeurIPS}, 2014.

\bibitem{vaswani2017attention}
Ashish Vaswani, Noam Shazeer, Niki Parmar, Jakob Uszkoreit, Llion Jones,
  Aidan~N Gomez, {\L}ukasz Kaiser, and Illia Polosukhin.
\newblock Attention is all you need.
\newblock In {\em NeurIPS}, 2017.

\bibitem{vedantam2015cider}
Ramakrishna Vedantam, C Lawrence~Zitnick, and Devi Parikh.
\newblock Cider: Consensus-based image description evaluation.
\newblock In {\em CVPR}, 2015.

\bibitem{Vinyals14}
Oriol Vinyals, Alexander Toshev, Samy Bengio, and Dumitru Erhan.
\newblock Show and tell: A neural image caption generator.
\newblock In {\em CVPR}, 2015.

\bibitem{Xu:ICML15}
Kelvin Xu, Jimmy Ba, Ryan Kiros, Kyunghyun Cho, Aaron Courville, Ruslan
  Salakhudinov, Rich Zemel, and Yoshua Bengio.
\newblock Show, attend and tell: Neural image caption generation with visual
  attention.
\newblock In {\em ICML}, 2015.

\bibitem{yang2021auto}
Xu Yang, Chongyang Gao, Hanwang Zhang, and Jianfei Cai.
\newblock Auto-parsing network for image captioning and visual question
  answering.
\newblock In {\em ICCV}, 2021.

\bibitem{Yang:CVPR19}
Xu Yang, Kaihua Tang, Hanwang Zhang, and Jianfei Cai.
\newblock Auto-encoding scene graphs for image captioning.
\newblock In {\em CVPR}, 2019.

\bibitem{yao2017incorporating}
Ting Yao, Yingwei Pan, Yehao Li, and Tao Mei.
\newblock Incorporating copying mechanism in image captioning for learning
  novel objects.
\newblock In {\em CVPR}, 2017.

\bibitem{yao2018exploring}
Ting Yao, Yingwei Pan, Yehao Li, and Tao Mei.
\newblock Exploring visual relationship for image captioning.
\newblock In {\em ECCV}, 2018.

\bibitem{yao2019hierarchy}
Ting Yao, Yingwei Pan, Yehao Li, and Tao Mei.
\newblock Hierarchy parsing for image captioning.
\newblock In {\em ICCV}, 2019.

\bibitem{yao2017boosting}
Ting Yao, Yingwei Pan, Yehao Li, Zhaofan Qiu, and Tao Mei.
\newblock Boosting image captioning with attributes.
\newblock In {\em ICCV}, 2017.

\bibitem{You:CVPR16}
Quanzeng You, Hailin Jin, Zhaowen Wang, Chen Fang, and Jiebo Luo.
\newblock Image captioning with semantic attention.
\newblock In {\em CVPR}, 2016.

\bibitem{zhou2021semi}
Yuanen Zhou, Yong Zhang, Zhenzhen Hu, and Meng Wang.
\newblock Semi-autoregressive transformer for image captioning.
\newblock In {\em ICCV}, 2021.

\end{thebibliography}
}

\end{document}